\documentclass{article}

\usepackage[final,nonatbib]{midl_2018}

\usepackage[utf8]{inputenc} 
\usepackage[T1]{fontenc}    
\usepackage{hyperref}       
\usepackage{url}            
\usepackage{booktabs}       
\usepackage{amsfonts}       
\usepackage{nicefrac}       
\usepackage{microtype}      
\usepackage{graphicx}
\usepackage{float}
\usepackage{amsmath}

\usepackage{wrapfig,lipsum,booktabs}
\usepackage{subcaption}

\title{A Deep Learning Framework for Automatic Diagnosis in Lung Cancer}

\author{
  Nikolay Burlutskiy\thanks{Corresponding author: Nikolay Burlutskiy <nikolay.burlutsky@contextvision.se>.}, Feng Gu, Lena Kajland Wilen \\
  ContextVision AB\\
  Linkoping, Sweden \\
  \texttt{\{nikolay.burlutsky, feng.gu, lena.kajland\}@contextvision.se} \\
  \And
  Max Backman, Patrick Micke \\
  Department of Immunology, Genetics and Pathology \\
  Uppsala University, Uppsala, Sweden \\
  \texttt{\{max.backman, patrick.micke\}@igp.uu.se} \\
}

\begin{document}

\maketitle

\begin{abstract}
We developed a deep learning framework that helps to automatically identify and segment lung cancer areas in patients' tissue specimens. The study was based on a cohort of lung cancer patients operated at the Uppsala University Hospital. The tissues were reviewed by lung pathologists and then the cores were compiled to tissue micro-arrays (TMAs). For experiments, hematoxylin-eosin stained slides from 712 patients were scanned and then manually annotated. Then these scans and annotations were used to train segmentation models of the developed framework. The performance of the developed deep learning framework was evaluated on fully annotated TMA cores from 178 patients reaching pixel-wise precision of 0.80 and recall of 0.86. Finally, publicly available Stanford TMA cores were used to demonstrate high performance of the framework qualitatively.
\end{abstract}

\section{Introduction}
The microscopic evaluation of cancer is the corner stone of clinical diagnostics, and particularly in lung cancer where the evaluation is highly dependent on the experience of a pathologist. The evaluation can vary considerably between individual pathologists. Digitalization of tissue specimens facilitates training of models for automatic diagnosis. Nevertheless, it is impossible to fill the entire image of a TMA into the VRAM of a modern GPU. Thus, the image must be split into patches treated as examples in a semantic segmentation problem. Then the predicted patches are stitched to form the entire TMA prediction. Therefore, we developed a deep learning framework to learn from such data.

\section{The Deep Learning Framework}
\label{deep_learning_framework}

Let $(\mathbf{x}_i, y_i)\subseteq\mathbf{X}\times\mathbf{Y}$ be a patch or an example of a given dataset, where $i\in\mathbb{N}$. The shapes are $h\times w\times 3$ and $h\times w\times 1$ respectively. We can simplify and formulate a deep network as a function $f(\mathbf{x};\theta)$, where $\theta$ is a collection of parameters of all the parametrized layers. The learning task is a process of searching for the optimal set of parameters $\hat{\theta}$ that minimizes a loss function $\mathcal{L}(y, f(\mathbf{x};\theta))$. A commonly used loss function for binary task (`benign' and `cancer') is sigmoid cross entropy loss. First, the output of the function $f$ is transformed to a probabilistic value in the range of $[0,1]$ via a sigmoid function as follows
\begin{equation}
\label{eqn:sigmoid_fun}
\sigma(z)= \frac{1}{1+\exp(-z)}
\end{equation}
Then the cross entropy loss is computed as
\begin{equation}
\label{eqn:loss_fun}
\mathcal{L}(y, f(\mathbf{x};\theta))=-y\log(\sigma(f(\mathbf{x};\theta))-(1-y)\log(1-\sigma(f(\mathbf{x};\theta)))
\end{equation}

Search of the optimal set of parameters $\hat{\theta}$ is known as optimization in machine learning. Popular optimizers include Stochastic Gradient Descent (SGD), Adaptive Gradient (AdaGrad), and Root Mean Square Propagation (RMSProp) \cite{deep_learning_book_2016}. Recently, Adaptive Moment Estimation (Adam) was proposed in \cite{kingma2015} that has become a particularly popular method for optimizing deep networks and is the optimizer used in our framework. 

\paragraph{Semantic Segmentation Networks} A semantic segmentation network predicts every pixel of a patch as one of the class labels. We selected $7$ most popular network architectures that are considered more capable than others for semantic segmentation such as `123S', `dilatednet', `drn-C26', `drn-C42', `unet', `densenet-D56', and `densenet-D103'.

The {\bf 123S} network is based on the classic FCN-8S \cite{long2015}. The network is relatively compact, and thus less prone to overfitting. The {\bf dilatednet} is a network derived from FCN-32S \cite{long2015}, where the transposed convolution layers are replaced by convolution layers with dilation filters \cite{yu2016}. The dilated residual networks ({\bf drn-C26/C42}) \cite{yu2017} combine dilated convolution and residual networks (ResNet) \cite{yu2015}, which employ residual units to build very deep networks for image classification. The {\bf unet} \cite{ronneberger2015} has an encoder for downsampling and a decoder for upsampling, which are linked to form a U shape and thus the name. The network has high capacity of feature representation, and it can learn and aggregate knowledge at multiple scales in the data. It has become one of the most popular networks for semantic segmentation problems in biomedical imaging. Finally, DenseNet is a very deep network architecture that uses `dense blocks' to allow each layer to connect every other layer in a feed-forward fashion. The FC-DenseNet \cite{jegou2016} replaces all the fully connected layers with fully convolution layers. There are 56 layer in {\bf densenet-D56} and 103 layers in {\bf densenet-D103} respectively.

\section{Experimental Setup and Results}
\label{experimental_conditions}

\begin{figure}
  \begin{subfigure}{.5\textwidth}
  \centering
  \includegraphics[width=0.95\textwidth,scale=0.8]{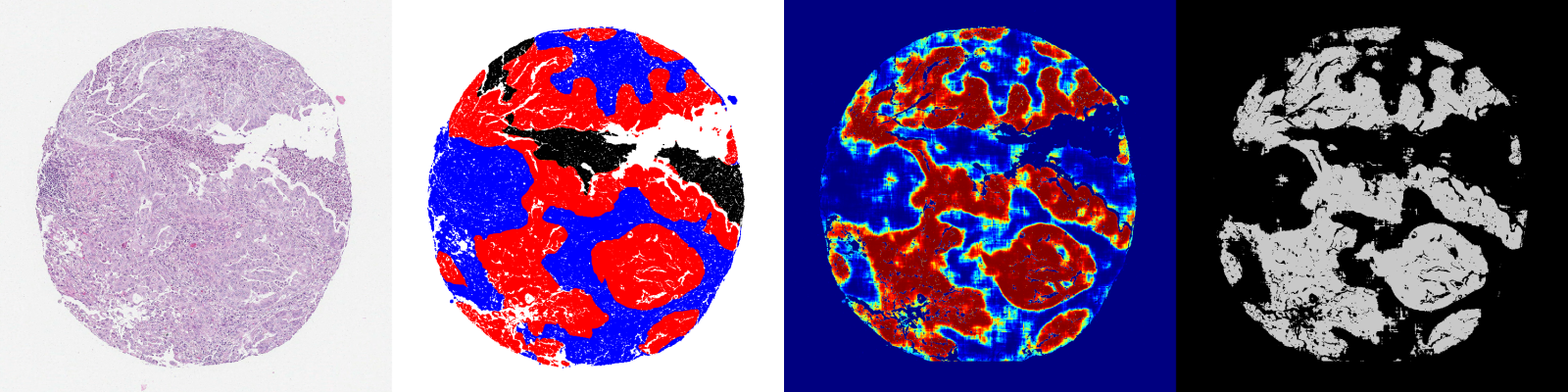}
  \includegraphics[width=0.95\textwidth,scale=0.8]{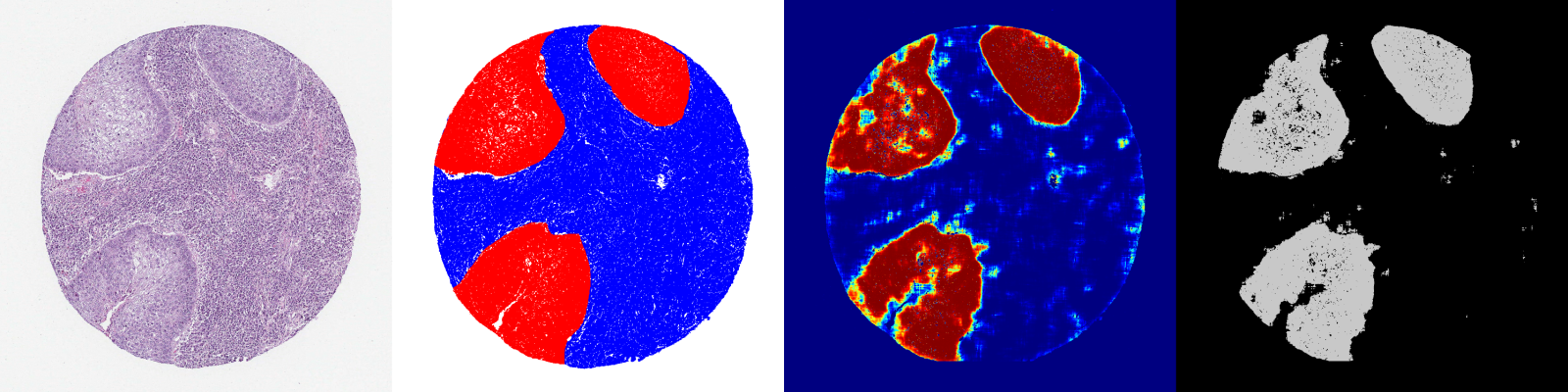}
  \caption{Examples of predicted cancer in 2 test cores.}
  \label{tma_predictions}
  \end{subfigure}%
  \begin{subfigure}{.5\textwidth}
  \centering
  \includegraphics[width=0.95\textwidth,scale=0.8]{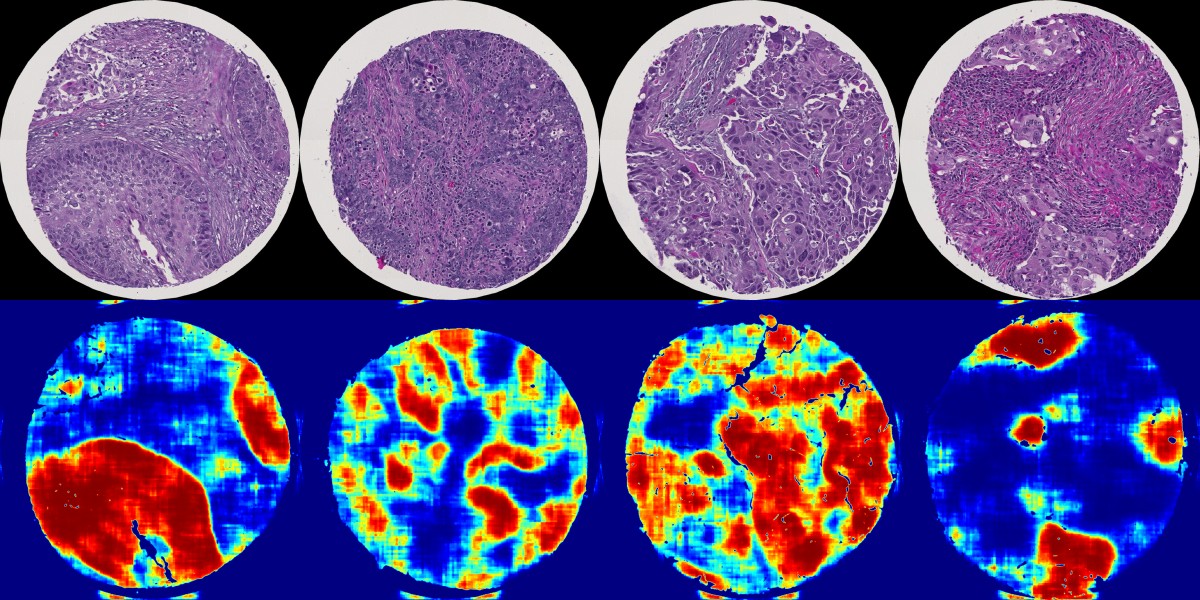}
  \caption{Examples of predicted cancer in 4 Stanford TMAs.}
  \label{stanford_tma_cores}
  \end{subfigure}%
  \caption{Examples of predicted cancer areas by the deep learning framework ({\bf{123S}} model).}
\end{figure}

The developed deep learning framework was trained and then evaluated on a lung cancer dataset. The dataset was formed from a cohort of 712 lung cancer patients operated in Uppsala Hospital including 223 squamous, 398 adenocarcinoma, 74 large cell cancer, and 17 other cancer types. For experiments, TMAs with H\&E stained cores were chosen. All cores were scanned as high resolution images at 0.5 $\mu m$/pixel, corresponding to approximately 9 million RGB pixels for each core. Then cancer (red), stroma (blue), necrosis (black), normal lung tissue areas (green), and areas to exclude from training (yellow) were annotated in the images (see Figure \ref{tma_predictions}, the second column with 2 annotated cores). The annotation was performed by seven experts including two specialist pathologists. 

\begin{wraptable}{r}{7cm}
  \caption{Evaluation Results for different models. }
  \label{evaluation_results}
  \begin{tabular}{lccc}
    \toprule
    \cmidrule{1-4}
    Model          & Precision & Recall & F1-score \\
    \midrule
    123S           & 0.80      & 0.85      & 0.80  \\
    dilatednet     & 0.80      & 0.83      & 0.79  \\
    drn-C26        & 0.78      & 0.85      & 0.79  \\
    drn-C42        & 0.79      & 0.85      & 0.80  \\
    \bf {unet}     & \bf{0.80} & \bf{0.86} & \bf{0.80}  \\
    densenet-D56   & 0.79      & 0.80      & 0.76  \\
    densenet-D103  & 0.79      & 0.79      & 0.76  \\
    \bottomrule
  \end{tabular}
\end{wraptable} 

The annotated $707$ cores were split into a training set of $354$ cores, a validation set of $175$ cores, and $178$ cores for a test set. These three datasets were balanced with respect to areas annotated as cancer, stroma, necrosis, and normal lung tissue areas. In total, 7 deep learning models for segmentation of cancer areas in the lung tissue were trained. The models were trained and evaluated on a cluster with 10 TitanXp GPUs with 12GB VRAM and 64GB RAM.

Firstly, the trained segmentation models were evaluated qualitatively by pathologists. The predictions demonstrated striking accuracy at pixel level (see Figure \ref{tma_predictions}). Then Precision, Recall, and F1-score metrics were calculated with {\bf{123S}} and {\bf{unet}} slightly outperforming other networks (see Table \ref{evaluation_results}, the metrics were calculated on the pixel level for each core and then averaged for 178 test cores). For evaluation, an optimal threshold for predictions was chosen in order to find a good trade-off between Precision and Recall of the models.

In our experiments, we predicted cancer heatmaps for a few cores from publicly available Stanford TMA cores \cite{StanfordTMA}. Pathologists qualitatively inspected these predictions and expressed an opinion of high potential of the trained models producing accurate predictions at the pixel level (see Figure \ref{stanford_tma_cores}). 

\section{Conclusions and Future Work}
\label{conclusions_and_future_work}

In this paper, we introduced a novel deep learning framework for accurate segmentation of cancerous areas in lung tissues. The framework demonstrated high pixel-wise performance in outlining cancer areas in lung TMA cores. The performance of the trained models was tested on publicly available Stanford TMAs as well. For this purpose, the predictions were examined qualitatively by trained pathologists. The trained models showed high potential in identifying cancer areas in TMA cores.

For future work, we plan to quantitatively evaluate the performance of the trained models on diverse lung tissues datasets. First, we will quantitatively evaluate the performance of our models on Stanford TMA cores by exhaustively annotating some of the cores. Also, we are going to test the trained models on biopsies and images from more clinical settings.

\subsubsection*{Acknowledgments}

We would like to thank Johanna Mattsson, Hedvig Elfving, Artur Mazheyeuski, Diana Djureinovic, Cecilia Lindskog and Sahar Sayegh for their help in annotations and providing pathology expertise for the project.  

\bibliographystyle{plain} 
\bibliography{bibliography} 

\end{document}